\documentclass{article}
\usepackage{spconf,amsmath,graphicx}

\title{Class LM and Word Mapping for Contextual Biasing in End-to-End ASR}
\name{Rongqing Huang, Ossama Abdel-hamid, Xinwei Li, Gunnar Evermann}
\address{Apple \\
\small \tt{\{huangr, oabdelhamid, xinwei\_li2, gevermann\}@apple.com}
}

\begin{document}
%
\maketitle
\setlength{\textfloatsep}{5pt}
\setlength{\intextsep}{5pt}
\begin{abstract}
In recent years, all-neural, end-to-end (E2E) ASR systems gained rapid interest in the speech recognition community. They convert speech input to text units in a single trainable Neural Network model. In ASR, many utterances contain rich named entities. Such named entities may be user or location specific and they are not seen during training. A single model makes it inflexible to utilize dynamic contextual information during inference. In this paper, we propose to train a context aware E2E model and allow the beam search to traverse into the context FST during inference. We also propose a simple method to adjust the cost discrepancy between the context FST and the base model. This algorithm is able to reduce the named entity utterance WER by 57\% with little accuracy degradation on regular utterances. Although an E2E model does not need a pronunciation dictionary, it’s interesting to make use of existing pronunciation knowledge to improve accuracy. In this paper, we propose an algorithm to map the rare entity words to common words via pronunciation and treat the mapped words as an alternative form to the original word during recognition. This algorithm further reduces the WER on the named entity utterances by another 31\%.
\end{abstract}
\section{Introduction}
\label{sec:intro}

Conventional speech recognition systems include several main components: acoustic model, language model, and pronunciation dictionary. Each of them is separately constructed and optimized. 
In recent years, an all-neural, end-to-end (E2E) model that directly converts speech into text through a sequence model became popular. Instead of separately optimized components, the E2E model is a single trainable neural network. It removes the HMM assumptions and enables end-to-end optimization. Architectures like Connectionist Temporal Classification (CTC) \cite{ref:graves14}, attention based sequence models such as Listen, Attend and Spell (LAS) \cite{ref:las15}, Recurrent Neural Network Transducer (RNN-T) \cite{ref:graves12} have obtained impressive results. In particular, architectures like LAS can sometimes outperform the conventional system \cite{ref:prabh17, ref:chiu18}. The base LAS model requires the whole input sequence before it can compute the attention. This makes it infeasible for online streaming application. In \cite{ref:raffel17, ref:mocha18}, a Monotonic Chunkwise Attention (MOCHA) was proposed for the attention based sequence-to-sequence model. The LAS-MOCHA structure becomes the base model for our study, although our proposed methods should be directly applicable to other types of E2E model.

Users' voice requests often involve personal content like contact names, app names, music titles, etc. There are a few issues here. First, such personal content frequently includes rare and foreign words. The general training set has very few occurrences of such words. Second, each user's personal content is different. The common entities may not be what the users want. For example, if a user has a contact \texttt{Jain Smith}, a phrase like \texttt{call Jain Smith} may be misrecognized as \texttt{call Jane Smith}. Another user may have a contact \texttt{Jaine Smith}. It's important to inject the user-specific content during inference. 

In conventional hybrid HMM-DNN system, the contextual information is usually represented as a Weighted Finite-State Transducer (WFST, \cite{ref:mohri08}), and injected into the main FST graph during the recognition \cite{ref:novak12, ref:aleksic15, lmepatent}.

A single E2E model lacks the flexibility to inject the contextual information during recognition. There have been various attempts to improve it, e.g. in \cite{ref:pundak18}, a separate attention component is added to model the contextual phrases (a.k.a. bias phrases); in \cite{ref:williams18, ref:zhao19, ref:chen19}, an on-the-fly rescoring (a.k.a. shallow fusion) with bias phrases was proposed. Our proposed method is similar to \cite{ref:williams18, ref:chen19} with a few important differences. We identify the bias phrases in the training data and model the transitions between regular words and bias words, while during inference, the user-specific bias phrases are inserted as WFST graphs at the relevant place in the beam search. We propose to normalize the scores between the paths from the contextual bias FST and from the base search space.
It's hard for the E2E system to output words it has never seen in the training data \cite{ref:bruguier19, ref:hu19}. Such words are especially prevalent in user contact lists and music lists. We propose a method to do word mapping, i.e, transform the rare words to common words through pronunciation. This method makes use of an existing lexicon and achieves nice accuracy gain. 


\section{Base LAS-MOCHA Model}
\label{sec:background}


Given the speech sequence $\textbf{x}=\left\{x_1,...,x_L\right\}$ with length $L$, and output word sequence $\textbf{y}=\left\{y_1,...,y_U\right\}$ with length $U$, the E2E model computes the probability
\begin{equation}
P(\textbf{y}|\textbf{x})=\prod_{i=1}^{U}P(y_i|\textbf{x},y_1,y_2,...,y_{i-1})
\end{equation}
An encoder converts $\textbf{x}$ to intermediate outputs $\textbf{h}=\left\{h_1,...,h_T\right\}$ through a Recurrent Neural Network (RNN), typically an LSTM. One important aspect for ASR is there are many more speech frames than the number of output tokens, usually there is a reduction factor $N$, thus $T=\frac{L}{N}$.
\begin{equation}
h_j = \mathrm{EncoderRNN}(x_j, h_{j-1})
\label{eq:enc}
\end{equation}
The decoder is also an RNN. It takes encoder outputs $\textbf{h}$ (a.k.a. memory), and previous output token $y_{i-1}$, generates the current decoder state $s_i$:
\begin{equation}
s_i = \mathrm{DecoderRNN}(y_{i-1}, s_{i-1}, c_i)
\label{eq:dec}
\end{equation}
which is then passed through a generation network, typically a feedforward network with softmax output, to produce the next output token $y_i$:
\begin{equation}
y_i=\mathrm{Generate}(s_i, c_i)
\end{equation}
$c_i$ is the vector at decoder step $i$ that summarizes information from the encoder:
\begin{equation}
c_i=\sum_{j=1}^{T}\beta_{i,j}h_j
\end{equation}
Where $\beta_{i,j}$ is attention weight at output step $i$ on $j$-th encoder output.
The difference between original LAS and the MOCHA attention is how $\beta_{i,j}$ is computed. MOCHA defines two levels of score functions for attention weight computation. Please refer to Section 2 in \cite{ref:mocha18} for details.  

\section{Context Injection}
\label{sec:lme}
\subsection{Context Aware Training}
Our LAS-MOCHA model takes 40-dimension mel-filter banks, and outputs BPE tokens (Byte Pair Encoding \cite{ref:sennrich2016}). With BPE tokens, we can in theory cover all the words in a language, so we don't have out-of-vocabulary (OOV) problems. This is important for the user-specific named entities. From the speech-text pair training data, we relabel the transcription to insert class LM tags. E.g, for utterance \texttt{call Jain Smith mobile}, it's relabelled into \texttt{call @contact\# Jain Smith \#contact@ mobile}; similarly for utterance \texttt{open ClassDojo}, it's relabelled into \texttt{open @app\# ClassDojo \#app@}. The tokens \texttt{@contact\#}, \texttt{@app\#}, \texttt{\#contact@}, \texttt{\#app@} are class enter and exit tokens for classes \texttt{contact} and \texttt{app} respectively. The class tags are excluded from BPE processing.
Human transcription is based on existing speech recognition system output that includes class LM tags. We apply an edit distance alignment between transcription and recognition output, then insert the class LM tags into the transcription. 
Alternatively, a process like named entity tagging can be used. Note that tagging does not need to be perfect as the content between the enter and exit tags will be replaced by user-specific FST during inference. 

\subsection{Contextual Bias FST Construction}
The contextual bias FST is a transducer from BPE subword sequence to word sequence (phrase). The cost on the FST arcs is derived from the relevant frequency $f_i$  for phrase $i$:
\begin{equation}
C_i=-\log\frac{f_i}{\sum_{j=1}^{N_l}f_j}
\end{equation}
where $N_l$ is the number of phrases in this context FST. If there are $M$ arcs on the path for this phrase, actual cost on each arc is 
\begin{equation}
C_{i}^{'}=\frac{C_i}{M}
\end{equation}
The transducer $T$ is determinized and minimized:
\begin{equation}
T_c = \mathrm{Min}(\mathrm{Det}(T))
\end{equation} 
We build one FST per class. Each use case (user-specific, location-specific, etc.) will have its own FSTs.
\subsection{Inference}
The contextual bias FST is constructed before inference. During inference, if an active token in the beam is one of the class enter tags like \texttt{@contact\#}, the corresponding context FST will be activated and its path will be traversed. Such path will compete against other paths inside the context FST, and also at the same time, compete against the paths in the base search space. When a final state of a context FST is reached, it traverses the class exit tag \texttt{\#contact@}, and gets back to the base search space. 
For $t$-th step of beam search, the score $S_{i}^{t}$ for $i$-th beam is:
\begin{equation}
S_{i}^{t}=\log P_{b}(y_{i,t}|y_{i,<t},\textbf{x})+\lambda _{c} \log P_{c}(y_{i,t}|y_{i,<t})
\label{eq:lme_score}
\end{equation}
where $P_{b}(\cdot)$ is probability from the base E2E model and $P_{c}(\cdot)$ is probability from the contextual bias FST, and $\lambda _{c}$ is the scale on the context FST. For the path outside of the context FST, the second term is 0, this makes the context FST path always worse than the base path (since $\log$ of probability is a negative number). Therefore, we define the following scoring function for paths outside of the context FST:
\begin{equation}
S_{i}^{t}=\log P_{b}(y_{i,t}|y_{i,<t},\textbf{x})+\lambda _{b} \Gamma _{t}
\label{eq:nonlme_score}
\end{equation}
Where $\lambda _{b}$ is a scale, and $\Gamma _{t}$ is the normalization score at step $t$:
\begin{equation}
\Gamma _{t}= \frac{1}{\kappa _{t}}\sum \limits_{i=1}^{\kappa _{t}}\log P_{c}(y_{i,t}|y_{i,<t})
\label{eq:nonlme_score_def}
\end{equation}
Where $\kappa _{t}$ is the number of active paths from the context FSTs at step $t$. If all the paths at step $t$ are in the base search space, $\Gamma _{t}=0$; if any paths at step $t$ are inside a context FST, Eq. \ref{eq:lme_score} defines $S_{i}^{t}$ for paths inside the context FST, and Eq. \ref{eq:nonlme_score} defines $S_{i}^{t}$ for paths in the base search space. Note that the actual ranking of the paths at step $t$ are based on the accumulated score:
\begin{equation}
S_{i}^{1,...,t}=\sum \limits_{k=1}^{t}S_{i}^{k}
\end{equation}

\section{Word Mapping Through Pronunciation}
\label{sec:wordmap}
The E2E system can output graphemes or words, therefore it does not need a pronunciation dictionary. On the other hand, we have created a large pronunciation dictionary in the many years of conventional system development. The grapheme-to-phoneme (G2P) system is also mature. It is an interesting research question how to inject such knowledge into the E2E system. Rare or even foreign words are common in the named entity phrases. The E2E system has difficulty generating words it rarely sees. In this section, we describe a method to convert the rare named entity words into more common words through pronunciation so they are easier to recognize.

Given a word n-gram model $G$ trained from the text data used in the E2E model training, and a pronunciation dictionary (lexicon) $L$, we construct
\begin{equation}
D=L \circ G
\end{equation}   
For a word $W$, its pronunciation is represented as a phoneme FST $P_W$, the pronunciation is either from a human generated lexicon or a G2P system and it may have multiple pronunciations for $W$ (the rare word). A new word $W'$ (the common word) is obtained through
\begin{equation}
W' = \mathrm{TopSort}(\mathrm{ShortestPath}(P_W \circ D))
\label{eq:pron_map}
\end{equation}
Then the subword list corresponding to $W'$ is used as the subword list for $W$ for contextual bias FST construction. We will also show experiments that adding the list to $W$ instead of replacing the original one.     
\section{Experiments}
\label{sec:exp}
We use English data in all our experiments. The base E2E model is LAS-MOCHA. The encoder is 5-layer unidirectional LSTM, 1400 cells, with 700-dim projections. 
The 40-dim mel-filter bank is the input to the encoder. 4-head, 800-dim, chunk-size 2 MOCHA attention is used. The decoder uses 2-layer unidirectional LSTM, each has 800 cells, and 400-dim projection. The text is processed with 6.4K BPE tokens.   
We use two test sets, first is a named entity rich set (\textit{named entity set}) with 13K utterances. Sample utterances include ``\texttt{call Jain Smith mobile}'' and ``\texttt{open ClassDojo}''; the other has no named entities (\textit{regular set}) and has 21K utterances. 
Note that the user-specific context FSTs are loaded even for the \textit{regular set} so this is a good test set for measuring the ``anti-biasing'' phenomenon.  Both test sets are typical voice assistant requests.

The E2E model is trained with block-momentum SGD \cite{chen2016scalable}, 32 V100 GPUs, 20000 frames per minibatch, L2 normalization on gradient, and cross-entropy criterion, with label smoothing \cite{ref:szegedy16} weight 0.05 and schedule sampling \cite{ref:bengio15} weight 0.1.  The initial learning rate is 0.025 and decayed by factor 0.8 when no improvement observed on the validation set. The model was trained for 25 epochs.
We use beam size 8 during beam search, and length penalty \cite{ref:wu16} 0.1. 

\subsection{Contextual bias FST in inference}
\label{sec:expt_inference}
The first set of experiments are to confirm the effectiveness of the proposed context FST algorithm in the inference. The amount of training data for the E2E model is a few thousand hours here. 

\begin{table}[th]
\caption{Effect of adding class LM tags.}
\label{tab:class_lm}
\centering
\begin{tabular}{|l|c|c|}
\hline
\textbf{Setup} & \textbf{Named entity set} & \textbf{Regular set} \\
& \textbf{WER} & \textbf{WER} \\
\hline
LAS-MOCHA & 19.2 & 9.1 \\
\hline
+class LM tags & 18.1 & 8.4 \\
\hline
\end{tabular}
\end{table}
%
The class LM enter-exit tags can help the model learn the transition between regular words and named entity words. This is evident from the results in Table \ref{tab:class_lm}, where only the class LM tags were added to the training text and the accuracy is already improved for both named entity test set and regular test set. 

Next, the contextual bias FST is added in the inference. When $\lambda _{c}=0$, it means no context FST is used; when $\lambda _{b}=0$, it means no score normalization is used. Table \ref{tab:score_norm} shows the impact of using context FSTs and score normalization. From setup 1 to 2, using context FST and scale 0.1 reduces the WER from 18.1 to 14.4; from setup 2 to 3, turning on score normalization alone reduce the WER from 14.4 to 11, changing the context FST scale from 1.0 to 0.1 further reduces the WER to 9.0. The overall WER reduction of using context FST is 50\% (from 18.1 to 9.0). The last column shows the WER on the \textit{regular set}, the WER is increased slightly from 8.4 to 8.5.  
\begin{table}[th]
\caption{Score normalization of contextual bias FST and its impact on accuracy.}
\label{tab:score_norm}
\centering
\begin{tabular}{|l|c|c|c|c|}
\hline
\textbf{ID} & $\lambda _{c}$ & $\lambda _{b}$ & \textbf{Named entity set} & \textbf{Regular set} \\
& & & \textbf{WER} & \textbf{WER} \\
\hline
1  & 0 & 0 & 18.1 & 8.4 \\
\hline
2 &  0.1 & 0 & 14.4 & 8.5 \\
\hline
3 & 1.0 & 1.0 & 11.0 & 8.5 \\
\hline
4 & 0.1 & 1.0 & 9.0 & 8.5 \\
\hline
\end{tabular}
\end{table}
%
\subsection{Word mapping through pronunciation}
\label{sec:word_map}
The E2E model used in this experiment is trained with four times more data, SpecAugment \cite{ref:park19}, and Minimum Word Error Rate criterion (MWER, \cite{prabhavalkar2018, shen2016}), the weight on the cross-entropy loss during MWER is 0.05. We ran one epoch of MWER after the cross-entropy training. The model parameters and other hyper-parameters are the same as in Section \ref{sec:expt_inference}. For word mapping, the unigram trained on the text portion of the training data is used and the base lexicon has less than a million unique words. A bi-LSTM based G2P was used to generate pronunciations for the OOV words in the user's context lists. The results are shown in Table \ref{tab:mapped_fst}.
    
\begin{figure}[t]
\centering
\includegraphics[scale=1,width=\linewidth]{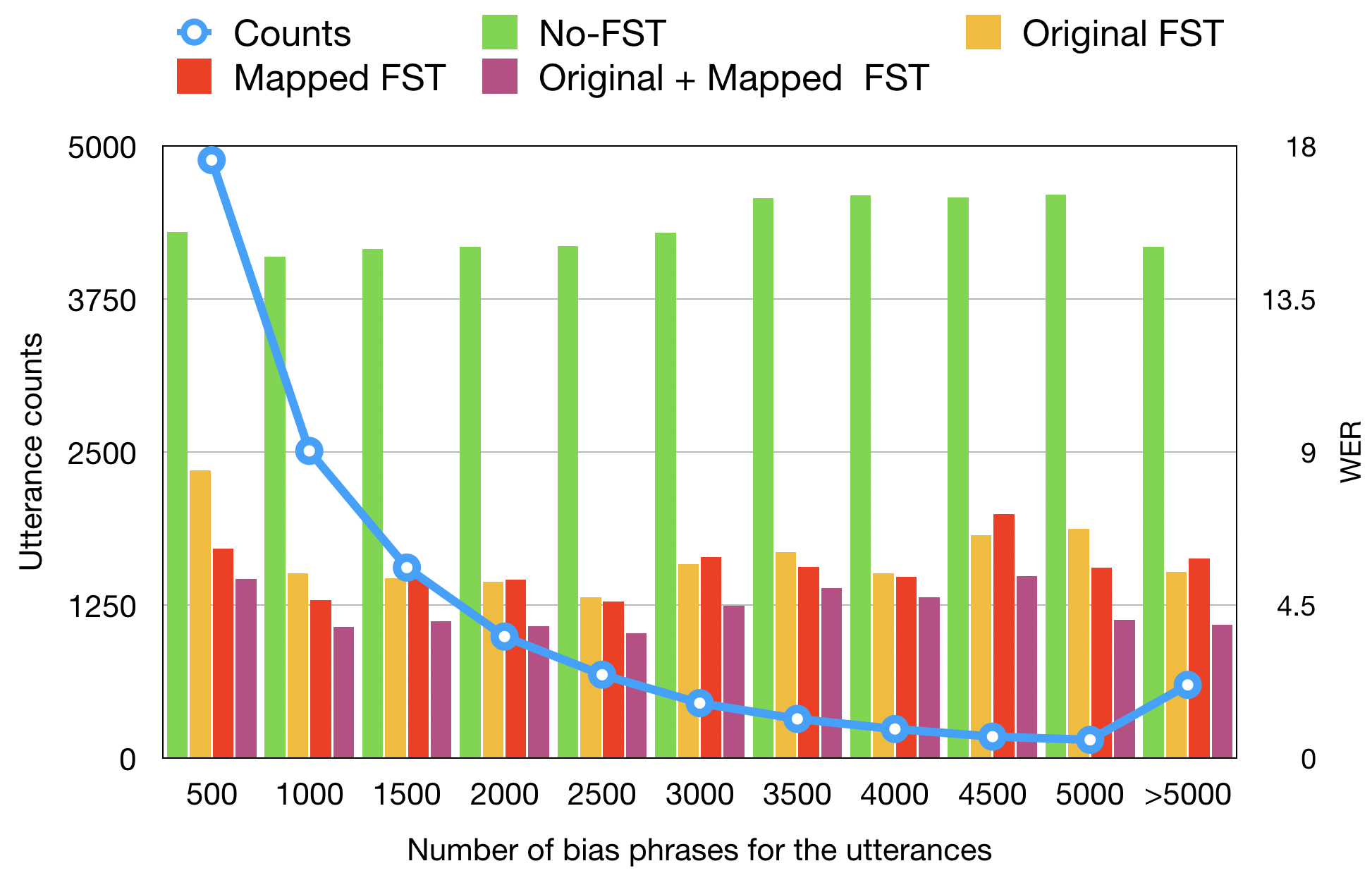}
\caption{Number of bias phrases in utterances and the WER breakdown by utterance buckets and configurations.}
\label{fig:bin_phrases}
\end{figure}

\begin{table}[th]
\caption{Word mapping through pronunciation and its impact on the accuracy.}
\label{tab:mapped_fst}
\centering
\begin{tabular}{|l|c|c|}
\hline
\textbf{Setup} & \textbf{Named entity set} & \textbf{Regular set} \\
& \textbf{WER} & \textbf{WER} \\
\hline
No context FST & 15.3 & 5.1 \\
\hline
FST from original words & 6.5 & 5.2 \\
\hline
FST from mapped words & 5.5 & 5.2 \\
\hline
Original+mapped FST & 4.5 & 5.2 \\
\hline
\end{tabular}
\end{table}
We use the configuration from setup 4 in Table \ref{tab:score_norm} for the context FST. The WER reduction from context FST is still big (from 15.3 to 6.5, a 57.5\% reduction). When using the subword sequence from the mapped words, the WER is reduced to 5.5. Interestingly, if using both the original sequence and mapped sequence in the FST, the WER is further reduced to 4.5. That's about 31\% error reduction over original context FST. The overall WERR over the no-FST baseline is \textbf{70.6\%}, with total inference time increased by 10\%. Fig. \ref{fig:bin_phrases}  shows the WER breakdown in terms of number of bias phrases for the utterances. The WER reduction is consistent across utterances with different number of bias phrases.

We examine some mapped words in Table \ref{tab:mapped_words}, the first column is the original word surface form, the second column is the mapped-to words through pronunciation, the third column is the recognized words when the original context FST is used. After using the subword sequence of the mapped word in the FST, these cases are fixed. From Table \ref{tab:mapped_words}, we observe that  rare words are mapped to more common words, either one-to-one mapping, or the original word is decomposed into multiple more common words, or less frequently, a phrase is reduced to a single word. 
\begin{table}[th]
\caption{Sample words mapped through pronunciation. With the mapping, original words are correctly recognized.}
\label{tab:mapped_words}
\centering
\begin{tabular}{|l|l|l|}
\hline
\textbf{Original} & \textbf{Mapped-to} & \textbf{Original} \\
\textbf{word} & \textbf{word} & \textbf{recognition} \\
\hline
Yvanna & ivana & ivana \\
\hline
sista & sister & sister \\
\hline
Ellie Gershenwald & Elle Gershon walled & Ellie garcia walt \\
\hline
Vandendriessche & Vanden Drey Eske & vandendraci \\
\hline
La Juana & Lajuana & Louetta \\
\hline
\end{tabular}
\end{table}
     
\section{Conclusions}
\label{sec:conc} 
User-specific content or location dependent content are  challenging for E2E models since the model directly converts speech to text. In this study, we have empirically demonstrated that inserting class LM tags into the text for E2E model training is beneficial by itself. The contextual bias FST is a useful technique to inject external knowledge into the E2E model. To make this technique more accurate, we have proposed a simple score normalization algorithm. The accuracy improvement is 57\%. We have also proposed to transform words through pronunciation. This algorithm converts the rare and unusual words into more common words so they are easier to recognize. This algorithm improves the accuracy by another 31\%. The overall accuracy improvement over the no-FST baseline is 70.6\%. Importantly, the proposed techniques cause little degradation on the non-named-entity type utterances.  In the future, we plan to incorporate this word mapping process into training so the model is more accustomed to the sequences from both the original word and mapped-to word.

\section{Acknowledgement}
We would like to thank Matt Mirsamadi, Kyuyeon Hwang, Tim Ng, Roger Hsiao, Henry Mason, Leo Liu, Arnab Ghoshal, Yuchen Zhang, Man-Hung Siu, and John Bridle for helpful discussions.


\vfill\pagebreak

\bibliographystyle{IEEEbib}
\bibliography{refs}

\end{document}